%% file: main.tex
\documentclass{ecai} 

\usepackage{latexsym}
\usepackage{amssymb}
\usepackage{amsmath}
\usepackage{amsthm}
\usepackage{booktabs}
\usepackage{enumitem}
\usepackage{graphicx}
\usepackage{color}
\usepackage{lipsum}
\usepackage{listings}
\usepackage{algorithm}
\usepackage{algpseudocode}
\usepackage{caption}

\newcommand{\BibTeX}{B\kern-.05em{\sc i\kern-.025em b}\kern-.08em\TeX}

\begin{document}


\begin{frontmatter}


\paperid{1654} 

\title{When to Accept Automated Predictions and When to Defer to Human Judgment?}

\author{\fnms{Daniel}~\snm{Sikar}\thanks{Corresponding Author. Email: daniel.sikar@city.ac.uk.}}
\author{\fnms{Artur}~\snm{Garcez}}
\author{\fnms{Tillman}~\snm{Weyde}}
\author{\fnms{Robin}~\snm{Bloomfield}}
\author{\fnms{Kaleem}~\snm{Peeroo}}

\address{Department of Computer Science, School of Science and Technology, City, University of London}

\input{Sections/01-Abstract}
\end{frontmatter}
\input{Sections/02-Introduction}

\input{Sections/03-Context}

\input{Sections/04-Methods}

\input{Sections/05-ResultsAndDiscussion}
\input{Sections/06-ConclusionsAndFutureWork}
\clearpage
\bibliography{references}
\end{document}

%% file: Sections/01-Abstract.tex


\begin{abstract}
Ensuring the reliability and safety of automated decision-making is crucial.
This paper proposes a new approach for measuring the reliability of predictions in machine learning models.
We analyze how the outputs of a trained neural network change using clustering to measure distances between outputs and class centroids.
We propose this distance as a metric to evaluate the confidence of predictions.
We assign each prediction to a cluster with centroid representing the mean softmax output for all correct predictions of a given class.
We then define a safety threshold for a class as the smallest distance from an incorrect prediction to the given class centroid.
We evaluate the approach on the MNIST and CIFAR-10 datasets using a Convolutional Neural Network and a Vision Transformer, respectively.
The results show that our approach is consistent across these data sets and network models, and indicate that the proposed metric can offer an efficient way of determining when automated predictions are acceptable and when they should be deferred to human operators.    
\end{abstract}

%% file: Sections/02-Introduction.tex

\section{Introduction}

Ensuring the reliability and safety of automated decision-making systems is important, particularly in high-impact areas where there exists potential for significant harm from errors \cite{amodei2016concrete}. Machine learning (ML) models, while powerful, are susceptible to making erroneous predictions when faced with data that differs from the distribution that they were trained on \cite{hendrycks2021many}. This phenomenon, known as distribution shift, poses a significant challenge in deploying ML in real-world scenarios \cite{quinonero2009dataset}.

Distribution shift is a pervasive issue in ML, occurring when the distribution of the data used to train a model differs from the distribution of the data that the model encounters during deployment \cite{quinonero2009dataset}. This discrepancy can lead to a significant degradation in model performance, as it may struggle to generalize to the new, unseen data \cite{hendrycks2019benchmarking}. 

Distribution shifts can manifest in various forms, such as covariate shift, concept drift, and domain shift \cite{moreno2012unifying}. Covariate shift arises when the input data distribution changes while the conditional distribution of the output given the input remains the same \cite{shimodaira2000improving}. Concept drift occurs when the relationship between the input and output variables changes over time \cite{gama2014survey}. Domain shift refers to the situation where the model is trained on data from one domain but applied to data from a different domain \cite{patel2015visual}.

To quantify and address distribution shift, researchers have developed various metrics and techniques. One common approach is to use statistical divergence measures, such as Kullback-Leibler (KL) divergence \cite{kullback1951information} or Maximum Mean Discrepancy (MMD) \cite{gretton2012kernel}, to assess differences in training and test data distributions. These metrics provide a quantitative understanding of the extent of the distribution shift.

Another approach is to employ domain adaptation techniques, which aim to align the feature distributions of the source and target domains \cite{wang2018deep}. This can be achieved through methods such as importance weighting \cite{sugiyama2007covariate}, feature transformation \cite{pan2009survey}, or adversarial learning \cite{ganin2016domain}. These techniques seek to mitigate the impact of distribution shift by making the model more robust to changes in the data distribution.

Recent work has also focused on developing algorithms that can detect and adapt to distribution shift in real-time \cite{lu2018learning}. These methods often rely on monitoring the model's performance on a stream of data and adjusting the model's parameters or architecture when a significant drop in performance is detected \cite{baena2006early}. Such adaptive approaches are particularly relevant in dynamic environments where the data distribution is likely to change over time.

Therefore, distribution shift is a significant challenge  that can lead to poor model performance if not addressed properly. 
As ML models are increasingly deployed in real-world applications, developing robust methods to handle distribution shift remains an important area of research.

In this paper, we propose a novel approach for quantifying the reliability of predictions made by neural networks under distribution shift. Our method leverages clustering techniques to measure the distances between the outputs of a trained neural network and class centroids in the softmax distance space. By analyzing these distances, we develop a metric that provides insight into the confidence that one may attribute to the model's predictions. We adopt the most conservative threshold value at which model predictions are expected to be 100\% accurate. The proposed metric offers a computationally efficient, practical way to determine when to accept an automated prediction and when human intervention may be necessary.

Furthermore, we explore the relationship between the distance to class centroids and the model's predictive accuracy. Our findings confirm as expected that classes predicted more accurately by the model tend to have lower softmax distances to their respective centroids. This observation highlights the potential for using the changes observed in the softmax distribution as a proxy for our confidence in the model's predictions. The objective is to establish a closer link than thus far identified by the literature between the distance-based metrics proposed for distribution shift and the approaches that measure the drop in model accuracy. It is expected to offer a better understanding of the connection between model performance and reliability, that is, accuracy and how confident we are in the predictions of neural networks. 



The main contributions of this paper are:
\begin{itemize}
\item A new lightweight approach for quantifying the reliability of neural networks under distribution shift using distance metrics and clustering to take full advantage of the learned softmax distributions.
\item A combined analysis of the metric-based and accuracy-based methods to tackle distribution shift, with experimental results showing consistency of the proposed approach across CNN and ViT architectures.
\end{itemize}


\subsection{Disclaimer}

While identifying and mitigating safety risks within the classification model itself is crucial, it does not guarantee the overall safety, robustness, reliability, trustworthiness, and confident performance of the entire system in which the model is deployed.

In critical applications, the classification model is often just one component of a larger, complex system. The safe and reliable operation of such a system depends on the proper functioning and interaction of all its components, including data acquisition, preprocessing, decision-making, and actuation. Even if the classification model is designed to mitigate safety risks, failures or vulnerabilities in other components can still compromise the overall safety and reliability of the system.

Furthermore, the deployment environment and context in which the classification system operates can introduce additional challenges and risks that may not be fully captured or addressed during the model development phase. Factors such as data distribution shifts, adversarial attacks, or unexpected edge cases can impact the model's performance and lead to potential safety issues.

Therefore, it is important to recognize that developing a safe and reliable classification model is necessary but not sufficient for ensuring the overall safety and reliability of the deployed system. A holistic approach that considers the entire system architecture, interfaces, and deployment context is essential. This includes rigorous testing, validation, and monitoring of the entire system, as well as establishing robust failsafe mechanisms and human oversight to handle unexpected situations and maintain safe operation.

In summary, while identifying and mitigating safety risks within the classification model is important, it is crucial to acknowledge that it does not guarantee the safe and reliable performance of the entire deployed system. A comprehensive approach considering all components, their interactions, and the deployment context is necessary to ensure the safe, robust, reliable, trustworthy, and confident operation of critical applications.

%% file: Sections/03-Context.tex

\section{Background}




\textbf{Softmax prediction probabilities}: The concept of using the entire set of softmax prediction probabilities, rather than solely relying on the maximum output, has been extensively studied in the context of enhancing the safety, robustness, and trustworthiness of machine learning models. By considering the complete distribution of class predictions provided by the softmax output, more reliable and informative prediction pipelines may be developed, that go beyond point estimates \cite{gal2016dropout}. This approach enables the exploration of uncertainty quantification, anomaly detection, and other techniques that contribute to building safer, more robust, and trustworthy autonomous systems.

Uncertainty quantification is a crucial aspect of reliable machine learning systems, as it allows for the estimation of confidence in the model's predictions \cite{kendall2017uncertainties}. By leveraging the softmax probabilities, techniques such as Monte Carlo dropout \cite{gal2016dropout} and ensembling \cite{lakshminarayanan2017simple} can be employed to estimate the model's uncertainty. These methods help identify instances where the model is less confident, enabling the system to defer to human judgment or take a more conservative action in safety-critical scenarios \cite{michelmore2018evaluating}.

Moreover, the softmax probabilities can be utilized for anomaly detection, which is essential for identifying out-of-distribution (OOD) samples or novel classes that the model has not encountered during training \cite{hendrycks17baseline}. By monitoring the softmax probabilities, thresholding techniques can be applied to detect anomalies based on the distribution of the predictions \cite{liang2018enhancing}. This enables the system to flag potentially problematic inputs and take appropriate actions, such as requesting human intervention or triggering fallback mechanisms.

The use of softmax probabilities also facilitates the development of more robust models that can handle adversarial examples and other types of input perturbations \cite{goodfellow2014explaining}. Adversarial attacks aim to fool the model by crafting input samples that lead to incorrect predictions with high confidence \cite{szegedy2013intriguing}. By considering the entire softmax distribution, defensive techniques such as adversarial training \cite{madry2017towards} and input transformations \cite{guo2018countering} can be applied to improve the model's robustness against these attacks.

Furthermore, the softmax probabilities provide valuable information for interpretability and explanability of the model's decisions \cite{ribeiro2016should}. By analyzing the distribution of the predictions, insights can be gained into the model's reasoning process and the factors that contribute to its outputs. This transparency is crucial for building trust in the system and facilitating human-machine collaboration \cite{doshi2017towards}.

The importance of leveraging the entire softmax distribution extends to various domains, including autonomous vehicles \cite{michelmore2018evaluating}, medical diagnosis \cite{leibig2017leveraging}, and financial risk assessment \cite{feng2018deep}. In these safety-critical applications, the consequences of incorrect predictions can be severe, and relying solely on the maximum softmax output may not provide sufficient safeguards. By considering the full distribution of predictions, more informed and reliable decisions can be made, reducing the risk of catastrophic failures.

However, the use of softmax probabilities is not without challenges. The calibration of the model's predictions is an important consideration, as poorly calibrated models may lead to overconfident or underconfident estimates \cite{guo2017calibration}. Techniques such as temperature scaling \cite{guo2017calibration} and isotonic regression \cite{zadrozny2002transforming} can be applied to improve the calibration of the softmax probabilities, ensuring that they accurately reflect the model's uncertainty.

By considering the complete distribution of class predictions, techniques can be employed to develop more reliable and informative prediction pipelines. 


\textbf{Clustering}: Clustering algorithms are essential for discovering structures and patterns in data across various domains \citep{jain2010data, xu2015comprehensive}. K-means, a widely used algorithm, efficiently assigns data points to the nearest centroid and updates centroids iteratively \citep{lloyd1982least}. However, it requires specifying the number of clusters and is sensitive to initial centroid placement \citep{arthur2007k}. Hierarchical clustering creates a tree-like structure by merging or dividing clusters \citep{johnson1967hierarchical} but may not scale well to large datasets \citep{mullner2011modern}. Density-based algorithms, like DBSCAN, identify clusters as dense regions separated by lower density areas \citep{ester1996density, schubert2017dbscan}.

Other applications include image segmentation for object detection \citep{shi2000normalized}, anomaly detection for fraud and intrusion detection \citep{chandola2009anomaly}, customer segmentation for targeted marketing \citep{ngai2009application}, and bioinformatics for gene expression analysis and disease subtype identification \citep{eisen1998cluster, jiang2004cluster}. The choice of algorithm depends on data characteristics, desired cluster properties, and computational resources \citep{rodriguez2019clustering}.

Given the context and to the best of our knowledge, no prior work exists in using softmax distance to class centroid to threshold trust in the model predictive accuracy in classification tasks.

%% file: Sections/04-Methods.tex

\section{Clustering and Softmax Distance as a Confidence Metric}


We consider a neural network output vector $\mathbf{p} = (p_1, p_2, \dots, p_K)$ where $\sum p_i = 1$, representing a probability distribution obtained by normalizing the logit vector $\mathbf{z} = (z_1, z_2, \dots, z_K)$ through the softmax function, $p_i = \text{softmax}(z_i) = e^{z_i} / \sum_{j=1}^{K} e^{z_j}$. For example, given $\mathbf{p} = [0.01, 0.01, 0.01, 0.01, 0.9, 0.01, 0.01, 0.01, 0.01, 0.01]$, the predicted class is '4', corresponding to the highest value at index five, reflecting the confidence of the prediction for each class from '0' to '9'. The logits, representing log-likelihoods of class memberships, are related to probabilities by $z_i = \log (p_i / (1 - p_i))$ where $z_i$ is the logit for class $i$, and $p_i$ is the probability of the input belonging to class $i$\cite{goodfellow2016deep, bishop2006pattern}.

We store the predictions for MNIST and CIFAR-10 datasets in a matrix $\mathbf{M} \in \mathbb{R}^{n \times 12}$, where $n$ is the number of predictions, the first ten columns are the softmax probabilities, column 11 is the true class and column 12 is the predicted class.

To obtain cluster centroids $\mathbf{C} \in \mathbb{R}^{10 \times 10}$ we calculate the mean of all correct predictions from the training datasets with Algorithm \ref{alg:k-means-centroid-init}. To calculate the softmax distance threshold we use all incorrect predictions with Algorithm \ref{alg:min_distance}.

\begin{algorithm}
\caption{K-Means Centroid Initialisation from Softmax Outputs}
\label{alg:k-means-centroid-init} 
\begin{algorithmic}[1]
\Require{$correct\_preds$: array of shape $(n, 12)$, where $n$ is the number of correct predictions}
\Ensure{$centroids$: array of shape $(10, 10)$, initialised centroids for each digit class}

\State $probs\_dist \gets corrects\_preds[:, :10]$ \Comment{Extract probability distribution for each digit}
\State $centroids \gets \text{zeros}((10, 10))$ \Comment{Initialise centroids array}

\For{$digit \gets 0$ to $9$}
\State $indices \gets \text{where}(\text{argmax}(probs\_dist, \text{axis}=1) == digit)[0]$ \Comment{Find indices of rows where digit has highest probability}
\State $centroid \gets \text{mean}(probs\_dist[indices], \text{axis}=0)$ \Comment{Compute mean probability distribution for selected rows}
\State $centroids[digit] \gets centroid$ \Comment{Assign centroid to corresponding row in centroids array}
\EndFor

\State \textbf{return} $centroids$
\end{algorithmic}
\end{algorithm}



\textbf{Convolutional Neural Network}: A Convolutional Neural Network (CNN) is used to classify handwritten digits from the MNIST dataset, consisting of 60,000 training images and 10,000 testing images, each of size 28x28 grayscale (single channel) pixels, representing digits from 0 to 9.

The CNN architecture, implemented using PyTorch, consists of two convolutional layers followed by two fully connected layers. The first convolutional layer has 16 filters with a kernel size of 3x3 and a padding of 1. The second convolutional layer has 32 filters with the same kernel size and padding. Each convolutional layer is followed by a ReLU activation function and a max-pooling layer with a pool size of 2x2. The output of the second convolutional layer is flattened and passed through two fully connected layers with 128 and 10 neurons, respectively. The final output is passed through a log-softmax function to obtain the predicted class probabilities.
\begin{algorithm}
\caption{Find Minimum Softmax Distances to Centroids for Incorrectly Predicted Digits (Threshold)}
\label{alg:min_distance} 
\begin{algorithmic}[1]
\Procedure{FindMinDistances}{$data$}
    \State $labels \gets [0, 1, 2, 3, 4, 5, 6, 7, 8, 9]$
    \State $thresh \gets \text{empty array of shape } (10, 2)$
    \For{$i \gets 0 \text{ to } 9$}
        \State $label \gets labels[i]$
        \State $min\_dist \gets \min(data[data[:, 1] == label, 0])$
        \State $thresh[i, 0] \gets min\_dist$
        \State $thresh[i, 1] \gets label$
    \EndFor
    \State \textbf{return} $thresh$
\EndProcedure
\end{algorithmic}
\end{algorithm}
The model was trained using the Stochastic Gradient Descent (SGD) optimizer with a learning rate of 0.01 and a batch size of 64. The learning rate was determined using a custom learning rate function that decreases the learning rate over time. The Cross-Entropy Loss function was used as the criterion for optimization. The model was trained for 10 epochs.

The MNIST dataset was preprocessed using a transformation pipeline that converted the images to PyTorch tensors and normalized the pixel values to have a mean of 0.5 and a standard deviation of 0.5. The dataset was then loaded using PyTorch's DataLoader, for batch processing and shuffling of the data.

The total number of parameters for the MNIST classification CNN is 206,922 and took 5m6s to train on a Dell Precision Tower 5810 with a 6 core Intel Xeon Processor and 32GB memory running Ubuntu 18.04.  

\textbf{Vision Transformer}: The Vision Transformer (ViT) architecture \cite{dosovitskiy2020image} is implemented using the Hugging Face Transformers \cite{wolf2020huggingfaces}
library. The model used in this study, 'google/vit-base-patch16-224-in21k' \cite{wu2020visual}, is pre-trained on the ImageNet-21k dataset \cite{deng2009imagenet}, which contains 14 million labeled images. The pre-trained model is then fine-tuned on the CIFAR-10 dataset.
The ViT model divides an input image into patches and processes them using a Transformer encoder. The model used in this study has a patch size of 16x16 and an image size of 224x224. The ViT model outputs a representation of the image, which is then passed through a linear layer to obtain the final class probabilities.
The model has 86.4M parameters and is fine-tuned to classify images from the CIFAR-10 dataset, consisting of 50,000 training images and 10,000 testing images, each of size 32x32 pixels, representing 10 different classes: airplane, automobile, bird, cat, deer, dog, frog, horse, ship, and truck.

The data preprocessing pipeline involves on-the-fly data augmentation using the torchvision library's transforms module. The training data undergoes random resized cropping, random horizontal flipping, conversion to a tensor, and normalization. The validation and testing data are resized, center-cropped, converted to a tensor, and normalized.

The model is trained using the Hugging Face Trainer API. The learning rate is set to $2 \times 10^{-5}$, the per-device train batch size is 10, the per-device eval batch size is 4, and the number of training epochs is 3. The weight decay is set to 0.01. The model is trained on Google Colab Pro, T4 GPU hardware accelerator with high ram.

%% file: Sections/05-ResultsAndDiscussion.tex

\section{Experimental Results and Discussion}

In this section, we present and discuss our results for CIFAR-10/ViT and MNIST/CNN. 

\begin{table*}[ht]
\centering
\begin{tabular}{c|cccccc|cccccc}
\hline
& \multicolumn{6}{c|}{Training Dataset Correct Predictions} & \multicolumn{6}{c}{Training Dataset Incorrect Predictions} \\
Digit & Mean & Median & Min & Max & $\sigma$ & Count & Mean & Median & Min & Max & $\sigma$ & Count \\ \hline
0 & 0.0166 & 0.0088 & 0.0018 & \textbf{0.8433} & 0.0536 & 5890 & 0.9955 & 0.9552 & \textbf{0.7037} & 1.3681 & 0.1756 & 33 \\
1 & 0.0154 & 0.0085 & 0.0020 & \textbf{0.7051} & 0.0486 & 6704 & 1.0912 & 1.0771 & \textbf{0.7247} & 1.3909 & 0.2197 & 38 \\
2 & 0.0370 & 0.0208 & 0.0056 & \textbf{0.7453} & 0.0746 & 5859 & 1.0409 & 1.0460 & \textbf{0.6962} & 1.3912 & 0.2064 & 99 \\
3 & 0.0352 & 0.0192 & 0.0055 & \textbf{0.7509} & 0.0754 & 6019 & 1.0522 & 1.0338 & \textbf{0.7139} & 1.3999 & 0.2143 & 112 \\
4 & 0.0295 & 0.0166 & 0.0031 & \textbf{0.6933} & 0.0661 & 5755 & 1.0681 & 1.0642 & \textbf{0.7139} & 1.4010 & 0.2045 & 87 \\
5 & 0.0298 & 0.0165 & 0.0024 & \textbf{0.7634} & 0.0693 & 5339 & 1.0572 & 1.0560 & \textbf{0.7066} & 1.3956 & 0.2000 & 82 \\
6 & 0.0154 & 0.0081 & 0.0015 & \textbf{0.7981} & 0.0485 & 5884 & 1.0485 & 1.0702 & \textbf{0.7068} & 1.4015 & 0.2110 & 34 \\
7 & 0.0302 & 0.0171 & 0.0028 & \textbf{0.7365} & 0.0678 & 6199 & 1.0133 & 0.9764 & \textbf{0.6960} & 1.3989 & 0.2160 & 66 \\
8 & 0.0619 & 0.0363 & 0.0092 & \textbf{0.7788} & 0.0989 & 5581 & 1.0232 & 1.0083 & \textbf{0.6852} & 1.3816 & 0.1990 & 270 \\
9 & 0.0399 & 0.0232 & 0.0048 & \textbf{0.7616} & 0.0781 & 5844 & 1.0576 & 1.0830 & \textbf{0.7179} & 1.3974 & 0.2201 & 105 \\ \hline
& \multicolumn{6}{c|}{Testing Dataset Correct Predictions} & \multicolumn{6}{c}{Testing Dataset Incorrect Predictions} \\
Digit & Mean & Median & Min & Max & $\sigma$ & Count & Mean & Median & Min & Max & $\sigma$ & Count \\ \hline
0 & 0.0141 & 0.0075 & 0.0021 & \textbf{0.7225} & 0.0506 & 977 & 1.0329 & 1.0185 & \textbf{0.7923} & 1.2880 & 0.2026 & 3 \\
1 & 0.0126 & 0.0071 & 0.0023 & \textbf{0.6737} & 0.0445 & 1129 & 0.9823 & 0.9556 & \textbf{0.7278} & 1.2445 & 0.1898 & 6 \\
2 & 0.0354 & 0.0202 & 0.0037 & \textbf{0.6697} & 0.0704 & 1015 & 1.0855 & 1.0725 & \textbf{0.7010} & 1.3833 & 0.1844 & 17 \\
3 & 0.0331 & 0.0177 & 0.0049 & \textbf{0.7515} & 0.0765 & 998 & 0.9823 & 0.9752 & \textbf{0.7102} & 1.3230 & 0.2044 & 12 \\
4 & 0.0306 & 0.0169 & 0.0047 & \textbf{0.6603} & 0.0678 & 972 & 0.9309 & 0.8725 & \textbf{0.7073} & 1.3110 & 0.1923 & 10 \\
5 & 0.0313 & 0.0173 & 0.0051 & \textbf{0.6324} & 0.0692 & 883 & 1.1874 & 1.2976 & \textbf{0.8477} & 1.3982 & 0.2067 & 9 \\
6 & 0.0172 & 0.0088 & 0.0029 & \textbf{0.7135} & 0.0545 & 941 & 1.0737 & 1.0242 & \textbf{0.7240} & 1.4009 & 0.2120 & 17 \\
7 & 0.0295 & 0.0165 & 0.0036 & \textbf{0.6939} & 0.0678 & 1009 & 0.9665 & 0.8833 & \textbf{0.7148} & 1.3628 & 0.2016 & 19 \\
8 & 0.0616 & 0.0372 & 0.0114 & \textbf{0.7328} & 0.0918 & 934 & 1.0513 & 1.0399 & \textbf{0.7252} & 1.3842 & 0.2022 & 40 \\
9 & 0.0359 & 0.0206 & 0.0042 & \textbf{0.6870} & 0.0765 & 980 & 1.0502 & 1.0486 & \textbf{0.6968} & 1.3838 & 0.2197 & 29 \\ \hline
\end{tabular}
\captionsetup{justification=raggedright,singlelinecheck=false}
\caption{Statistical Summary of MNIST Softmax Output Distances to Centroids for Different Datasets and Prediction Outcomes. The centroids are obtained from Softmax outputs for correct class predictions from the training dataset.}
\label{tab:distance_to_centroid}
\end{table*}

The CNN/MNIST classifier took 5m6s to train. The accuracy on the training dataset is 98.38\% and 98.46\% on the testing dataset. 
The ViT/CIFAR-10 classifier took 1h15m to train, registering 99.04\% accuracy on the testing dataset.

\noindent \textbf{Softmax Distance to Class Centroid Statistics:} We store the softmax outputs, true and predicted labels for the MNIST and CIFAR-10 training datasets, we obtain the mean values for all correctly predicted classes, use the values in algorithm \ref{alg:k-means-centroid-init}  to initialise the cluster centroids, then using the procedure described in algorithm \ref{alg:min_distance} and the same hardware used to train the MNIST classifier CNN, generate K-Means clusters and compute statistics as presented in Table \ref{tab:distance_to_centroid} (for the MNIST results), a summary of the distances between the softmax outputs of a model and the centroids for each digit class (0-9) in the MNIST dataset. The centroids are calculated using the softmax outputs for correct predictions from the training dataset. The table is divided into four main sections: Training Dataset Correct Predictions, Training Dataset Incorrect Predictions, Testing Dataset Correct Predictions, and Testing Dataset Incorrect Predictions. For each digit and section, statistics are displayed with respect to softmax distance from each class to its centroid: the mean, median, minimum (highlighted in bold for the testing datasets), maximum (highlighted in bold for the training datasets), standard deviation ($\sigma$), and count of instances are provided. We note that for the correct predictions, the mean softmax distance to class centroid for digits zero, one and six are lowest (network prediction accuracy is high) while for digit 8 the mean softmax distance to class centroid is a comparatively larger value (netrowk prediction accuracy is not as high). The significance of the bold highlight for Max values in the Training section, and Min values in the Correct Prediction sections, is that the Min values in the Incorrect Predictions sections, is that the latter is effectively the Threshold we propose be used as a boundary for accepting the class prediction as accurate. Anything softmax distances equal or above the threshold will be considered not safe to use. It becomes evident by comparing both columns, that for the training dataset, there is a significant overlap between Max and Min columns, however, the testing dataset presents an interesting property in that there is less of an overlap. The significance of the overlap is that any correct predictions within the overlap will be tagged as unsafe, for the scheme we propose to hold, which means more manual classification and human intervention on the downside, and on the upside a high likelihood that every prediction with softmax distance below the threshold is accurate.

In practice a small number of samples, 0.04\%, consistent across both CIFAR-10 and MNIST testing dataset prediction, are found in the overlap, as shown later in Tables \ref{tab:centroid_distance_overlap_mnist} and \ref{tab:centroid_distance_overlap}. Also shown in Table \ref{tab:distance_to_centroid} are the softmax distance standard deviations and counts for correct and incorrect classifications.

\begin{figure}[ht]
    \centering
    \includegraphics[width=0.99\columnwidth]{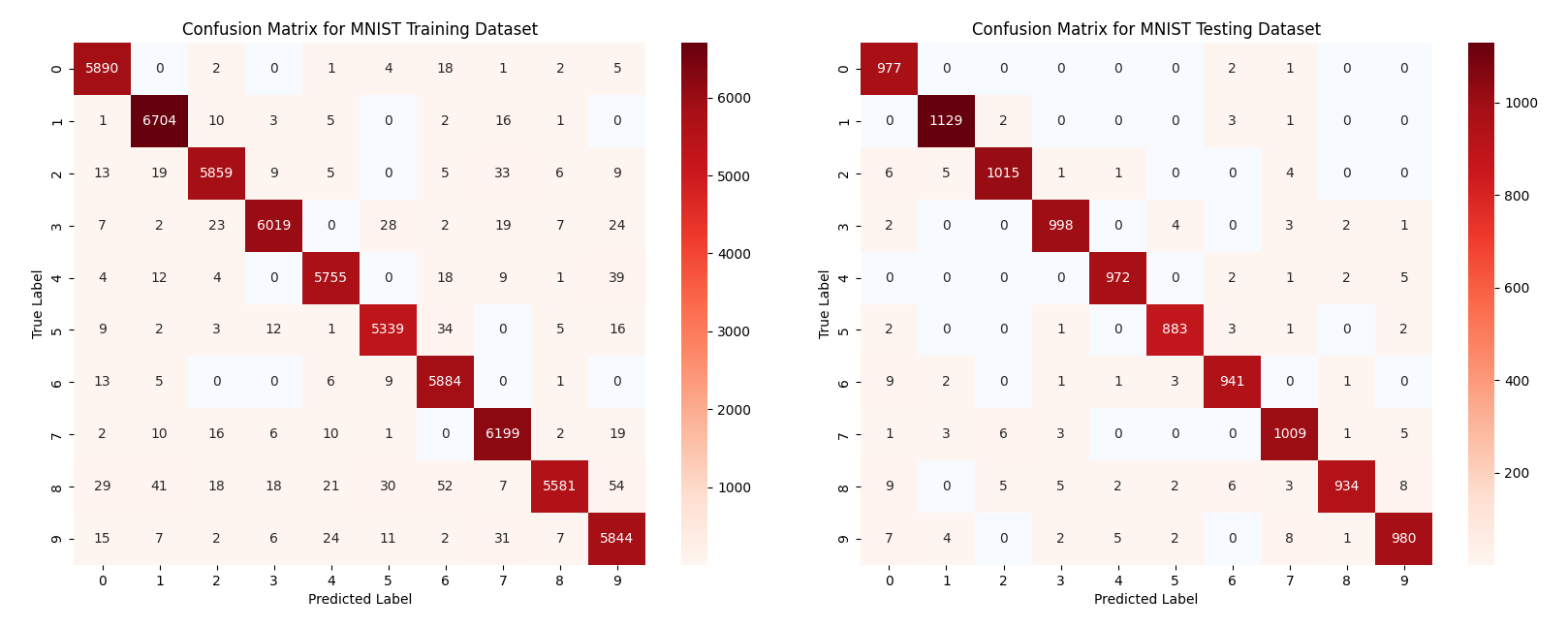}
    \caption{Please zoom in for detail. Confusion matrices for the MNIST classification model on the training dataset (left) and testing dataset (right). The matrices display the true labels on the vertical axis and the predicted labels on the horizontal axis. The diagonal elements represent correctly classified instances, while the off-diagonal elements indicate misclassifications.}
    \label{fig:mnist_combined_confusion_matrix}
\end{figure}

\noindent \textbf{Predictive Accuracy and Softmax Distance to Class Centroid:} The confusion matrices in Figure \ref{fig:mnist_combined_confusion_matrix} provide some insights into what are the most likely MNIST misclassifications. We note that number zero is the digit less likely, and number eight is the digit most likely to be misclassified. Other digits that are less likely to be misclassified are digits one and; six in the training dataset, and five in the testing dataset. Digit eight is likely to be misclassified as six or nine, while digit three is likely to be misclassified as two, five or nine.


\begin{figure}[ht]
    \centering   
    \includegraphics[width=0.99\columnwidth]{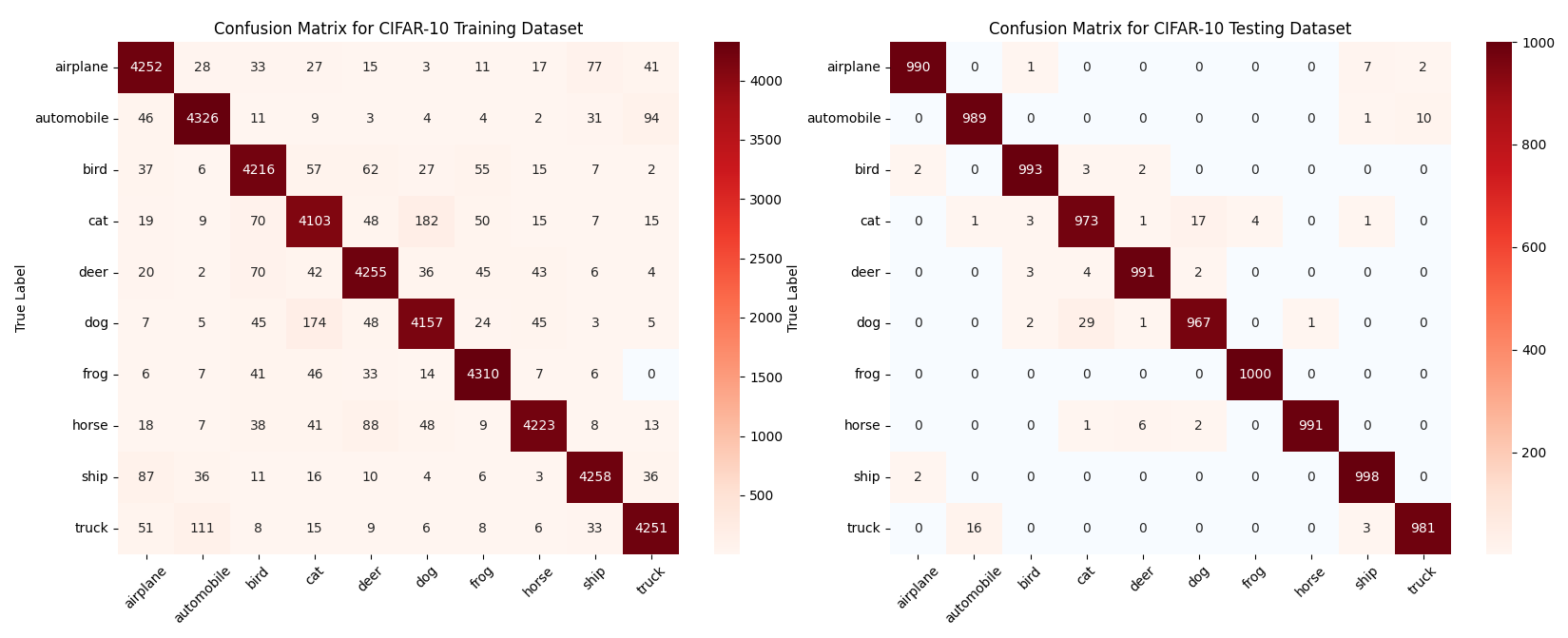}
    \caption{Please zoom in for detail. Confusion matrices for the CIFAR-10 classification model on the training dataset (left) and testing dataset (right).}
    \label{fig:cifar10_combined_confusion_matrix}
\end{figure}

The confusion matrices for the CIFAR-10 dataset in Figure \ref{fig:cifar10_combined_confusion_matrix} illustrate the performance of a classification model on both the training and testing sets. For the training set, the model exhibits a strong diagonal, indicative of high classification accuracy for most classes. Nonetheless, certain classes show notable confusion, such as 'cat' with 'dog' and 'deer' with 'frog', and the confusion is reflected in the class softmax distance to cluster centroid.

Examining the testing dataset, the model achieves good classification success, with perfect recognition of 'frog' class. Misclassifications are present but considerably reduced compared to the training dataset, particularly for pairs such as 'truck' and 'automobile', as well as 'cat' and 'dog'. The reduced confusion in the test set is also reflected in the softmax distance to cluster centroid.

\begin{figure}[ht]
    \centering
    \includegraphics[width=0.99\columnwidth]{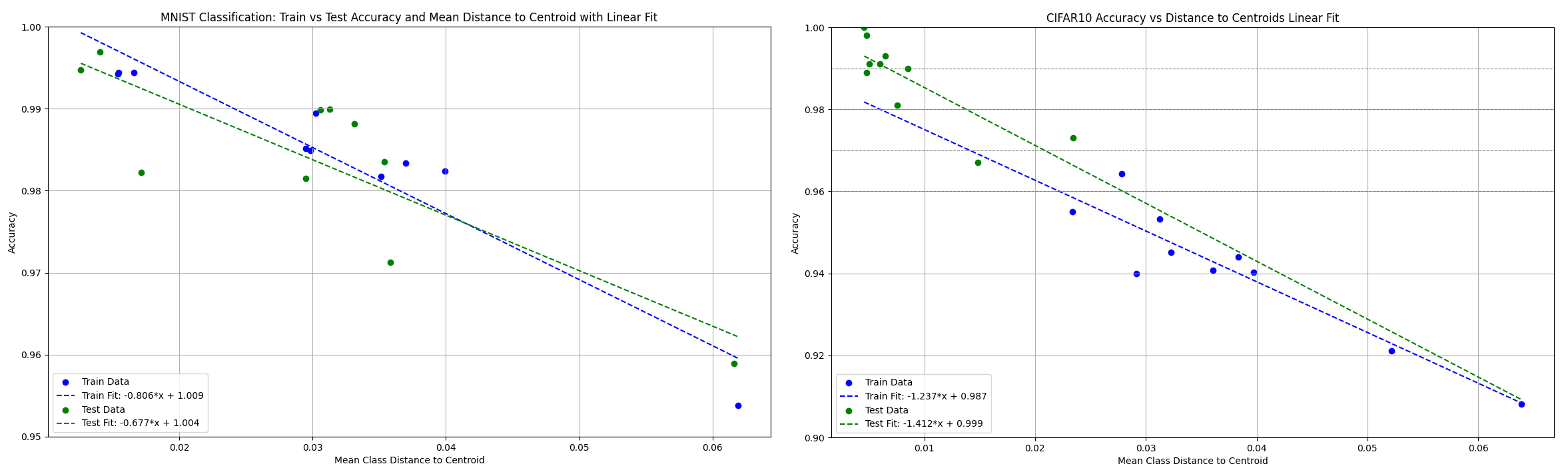}
    \caption{Please zoom in for detail. Expected accuracy linear fit based on prediction softmax distance to class centroid. MNIST fit is on the left, CIFAR-10 is on the right.}
\label{fig:Combined_CIFAR10_MNIST_Classification_Train_Test_Accuracy_Mean_Distance_to_Centroid_Linear_Fit}
\end{figure}

Figure \ref{fig:Combined_CIFAR10_MNIST_Classification_Train_Test_Accuracy_Mean_Distance_to_Centroid_Linear_Fit} suggests how the trained neural network model achieves higher predictive accuracy for classes that have lower softmax distance to their respective class centroids, compared to classes for which the model has lower predictive accuracy.


The scatter plot shows the relationship between classification accuracy and mean class distance to the centroid for the MNIST dataset. Training data (blue dots) and testing data (green dots) are each fitted with a linear regression line, demonstrating a negative correlation where increased mean distance corresponds to decreased accuracy. The steeper slope of the training data fit (-0.806) compared to the testing data fit (-0.677) reflects the greater accuracy and more compact cluster obtained from the correct test predictions compared to the cluster obtained from correct training predictions.

These results show an inverse relationship: as the mean distance from the data points in a class to their centroid increases, the propensity for correct classification diminishes. This could be due to the spread of data points within a class – greater distances from the centroid reflect a larger variance within the class, potentially making it more challenging for the classifier to identify the defining characteristics of each class accurately.

The CIFAR-10 linear fit is steeper as can be observed by the y axis values, although it can be argued provides a better fit.

\begin{figure*}[ht]
    \centering
    \includegraphics[width=0.99\textwidth]{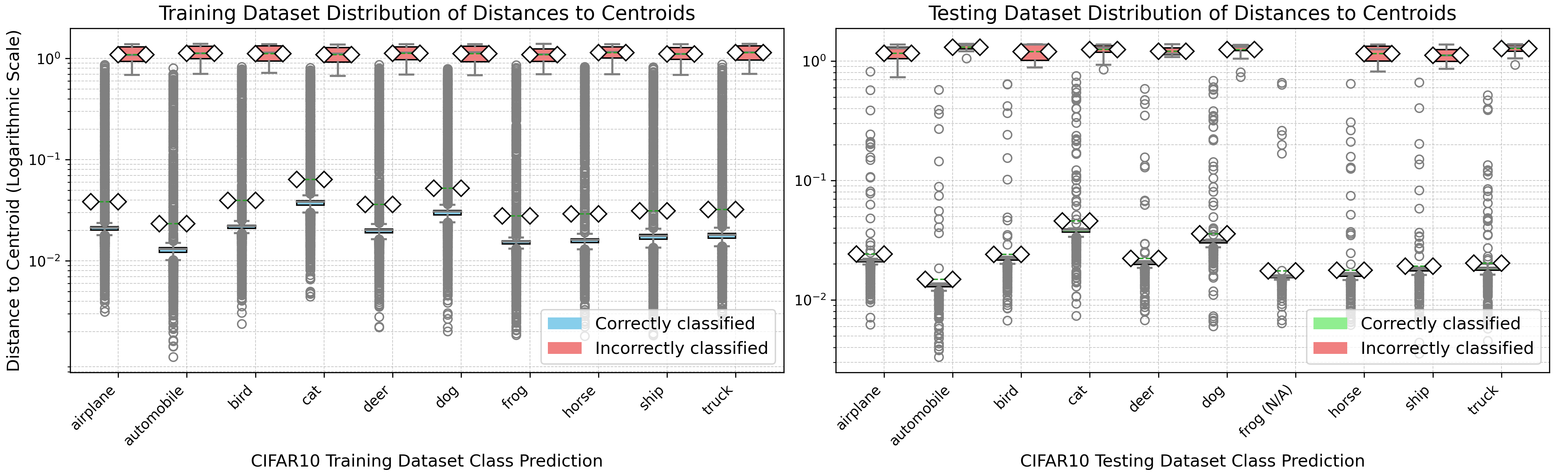}   \captionsetup{justification=raggedright,singlelinecheck=false}
    \caption{Distribution of Distances to Centroids for Correctly and Incorrectly Classified Instances in Training and Testing CIFAR-10 data, where training data is on the left and testing data is displayed on the right. Distances on y axis are shown on a logarithmic scale. Note, centroids are obtained from correctly classified training examples, then used for both training and testing datasets, a cluster is not created from the testing softmax distances.}
    \label{fig:CIFAR10_boxplots_side_by_side_x2}
\end{figure*}

\noindent \textbf{Overlap between maximum and minimum softmax distances to class centroids:} Figure \ref{fig:CIFAR10_boxplots_side_by_side_x2} presents box plots showing the distribution of distances to centroids for correctly and incorrectly classified instances in each class of the CIFAR-10 training and testing datasets. In the training dataset, the distances to centroids for correctly classified instances (green boxes) are generally lower than those for incorrectly classified instances (red boxes), indicating that correctly classified instances tend to be closer to their respective class centroids in the feature space. However, there is a small overlap between the distances of correctly and incorrectly classified instances, particularly in classes such as cat, dog, and deer. The testing dataset exhibits a similar trend, with correctly classified instances having lower distances to centroids compared to incorrectly classified instances. The overlap between the two groups is less pronounced in the testing dataset, such as in classes like frog. Note there is no frog boxplot for incorrect classifications, as reflected in the confusion matrix.

The side-by-side boxplots convey the distributions of distances to centroids for both the training and testing sets, categorized by class. These distances are represented on a logarithmic scale, facilitating the visualization of a wide range of values. For both sets, there is a clear pattern where the correctly classified instances tend to have a smaller median distance to the class centroid compared to the incorrectly classified ones. This observation is consistent across all classes, suggesting that instances closer to the centroid of their respective class in the feature space are more likely to be classified correctly by the model. The notable overlap in the interquartile ranges, however, indicates that while distance to the centroid is an informative metric, it is not solely determinative of classification accuracy.

The incorrectly classified instances exhibit greater variance in distances to the centroid, as evidenced by the longer whiskers and the presence of outliers. This variability implies that misclassified instances can sometimes be far off from the core cluster of their true class, potentially falling closer to the regions dominated by other classes in the feature space. This is indicative of boundary cases or instances that bear a resemblance to multiple categories. The consistency of these trends between the training and testing sets indicates that the distance to the centroid is a robust indicator of classification difficulty across the model's application. The presence of some correctly classified instances at higher distances suggests that other factors also influence classification accuracy, such as the density of data points in the surrounding feature space, class overlap, or specific intra-class variabilities.

\begin{table}[htbp]
\centering
\begin{tabular}{|c|c|c|c|c|c|c|}
\hline
Digit & \multicolumn{3}{c|}{Train Data - Train Centroids} & \multicolumn{3}{c|}{Test Data - Train Centroids} \\
\hline
 & Count & Total & Overlap & Count & Total & Overlap \\
\hline
0 & 2 & 5890 & 0.03\% & 1 & 977 & 0.10\% \\
1 & 0 & 6704 & 0.00\% & 0 & 1129 & 0.00\% \\
2 & 1 & 5859 & 0.02\% & 0 & 1015 & 0.00\% \\
3 & 3 & 6019 & 0.05\% & 1 & 998 & 0.10\% \\
4 & 0 & 5755 & 0.00\% & 0 & 972 & 0.00\% \\
5 & 5 & 5339 & 0.09\% & 0 & 883 & 0.00\% \\
6 & 2 & 5884 & 0.03\% & 1 & 941 & 0.11\% \\
7 & 2 & 6199 & 0.03\% & 0 & 1009 & 0.00\% \\
8 & 12 & 5581 & 0.22\% & 1 & 934 & 0.11\% \\
9 & 1 & 5844 & 0.02\% & 0 & 980 & 0.00\% \\
\hline
Totals & 28 & 59074 & 0.05\% & 4 & 9838 & 0.04\% \\
\hline
\end{tabular}
\caption{MNIST counts of correctly classified training and testing image predictions and counts with distance to centroids equal or above error threshold.}
\label{tab:centroid_distance_overlap_mnist}
\end{table}

\begin{table}[htbp]
\centering
\begin{tabular}{|c|c|c|c|c|c|c|}
\hline
Class & \multicolumn{3}{c|}{Train Data - Train Centroids} & \multicolumn{3}{c|}{Test Data - Train Centroids} \\
\hline
 & Count & Total & Overlap & Count & Total & Overlap \\
\hline
plane & 13 & 4252 & 0.31\% & 1 & 990 & 0.10\% \\
auto & 1 & 4326 & 0.02\% & 0 & 989 & 0.00\% \\
bird & 9 & 4216 & 0.21\% & 0 & 993 & 0.00\% \\
cat & 20 & 4103 & 0.49\% & 2 & 973 & 0.21\% \\
deer & 14 & 4255 & 0.33\% & 0 & 991 & 0.00\% \\
dog & 11 & 4157 & 0.26\% & 1 & 967 & 0.10\% \\
frog & 11 & 4310 & 0.26\% & 0 & 1000 & 0.00\% \\
horse & 12 & 4223 & 0.28\% & 0 & 991 & 0.00\% \\
ship & 11 & 4258 & 0.26\% & 0 & 998 & 0.00\% \\
truck & 10 & 4251 & 0.24\% & 0 & 981 & 0.00\% \\
\hline
Totals & 112 & 42351 & 0.26\% & 4 & 9873 & 0.04\% \\
\hline
\end{tabular}
\caption{CIFAR10 counts of correctly classified training and testing image predictions and counts with distance to centroids equal or above error threshold.}
\label{tab:centroid_distance_overlap}
\end{table}

Table \ref{tab:centroid_distance_overlap_mnist} presents an analysis of the counts and percentages of correctly classified images in the training and testing datasets that have softmax distances to their respective class centroids above a certain threshold. The threshold is determined by the nearest softmax distance to the centroid for the incorrectly classified classes.

The table is divided into two main sections: "Train Data - Train Centroids" and "Test Data - Train Centroids." For each section, the table provides the count of images that satisfy the threshold condition, the total number of images in each class, and the percentage of overlap (i.e., the proportion of images above the threshold relative to the total count).

Looking at the "Train Data - Train Centroids" section, we observe that the overlap percentages are relatively low, ranging from 0.00\% to 0.22\%. This suggests that only a small fraction of the correctly classified training images have softmax distances to their centroids above the threshold set by the incorrectly classified images. The highest overlap is observed for digit class 8, with 12 out of 5,581 images (0.22\%) having distances above the threshold.

Similarly, in the "Test Data - Train Centroids" section, the overlap percentages are even lower, ranging from 0.00\% to 0.11\%. This indicates that the trained model is able to correctly classify the majority of the test images while maintaining a distance to the centroids below the threshold set by the incorrectly classified images. The highest overlap in the test data is observed for digit classes 6 and 8, with 1 out of 941 (0.11\%) and 1 out of 934 (0.11\%) images, respectively, having distances above the threshold.

The Totals summarise the overall counts and overlap percentages across all digit classes. For the training data, 28 out of 59,074 images (0.05\%) have distances above the threshold, while for the test data, only 4 out of 9,838 images (0.04\%) exceed the threshold.

The low overlap percentages indicate that the majority of the correctly classified images are well-separated from the incorrectly classified images in terms of their softmax distances to the centroids. This analysis provides insights into the softmax distance being a proxy to distinguish between between correct and incorrect digit classifications based on the proximity of the images to their respective class centroids in the softmax distance space. The same analysis applies to Table \ref{tab:centroid_distance_overlap}. Most indicative of the consistency of the suggested approach is than both CIFAR-10 and MNIST present an overlap of 0.04\% for correctly classified images at or above threshold. Note we tag the images in the overlap as incorrectly classified, for the sake of assuring that all softmax distances to cluster centroid below threshold are correct classifications.

\begin{figure}[ht]
    \centering
    \includegraphics[width=0.99\columnwidth]{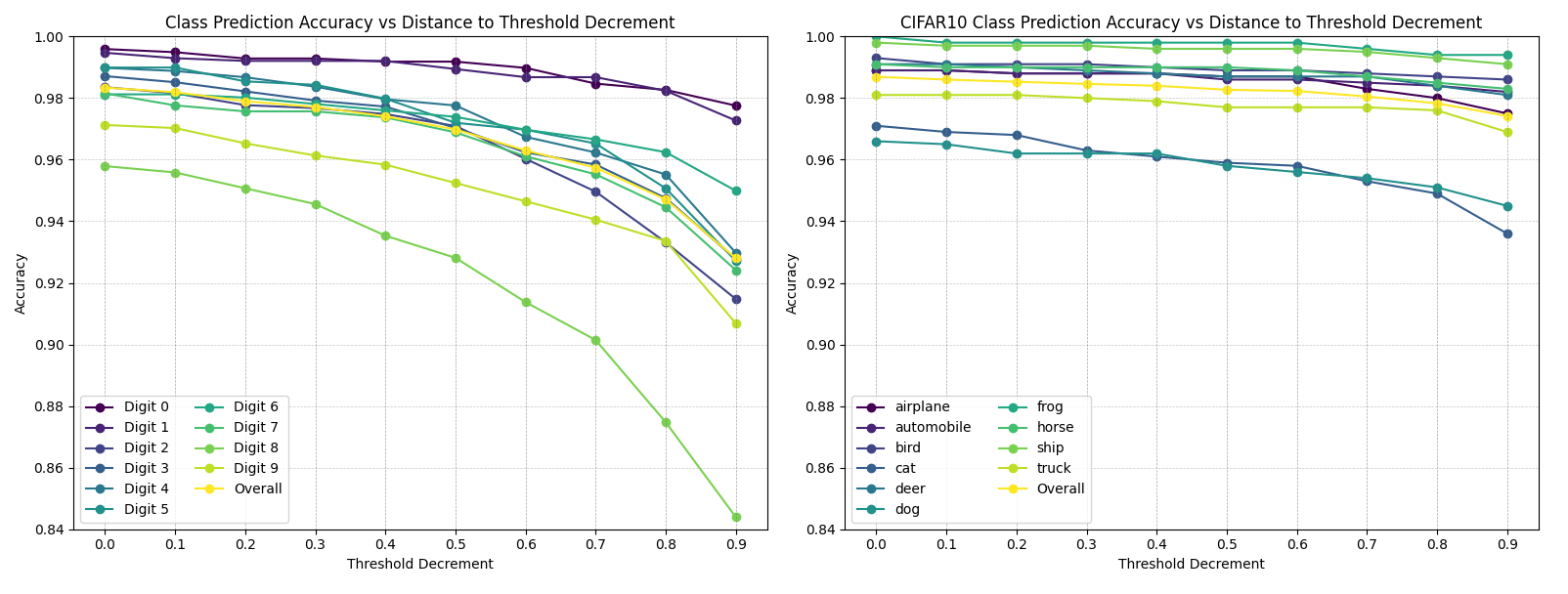}
    \caption{Please zoom in for detail. Expected accuracy decrease as a result of threshold decrease. MNIST data is on the left, CIFAR-10 is on the right. The x axis shows the threshold decrement in factors of 0.1, that is, at 0.1 the threshold is 90\% of the original threshold while at 0.9 the threshold is 10\% of the original threshold and consequently neared to the class centroid.}
\label{fig:Combined_CIFAR10_MNIST_single_plot_accuracy_decrements.png}
\end{figure}

\noindent \textbf{Accuracy decrease vs threshold decrease:} Having observed that the percentage of correctly classified images that have softmax distances equal or above threshold is 0.04\% for both CIFAR-10 and MNIST testing datasets, we ask how lowering the threshold affects the accuracy. Figure \ref{fig:Combined_CIFAR10_MNIST_single_plot_accuracy_decrements.png} reveals that for MNIST (left plot) classes accurately classified by the CNN model, such as zero and one, even setting the threshold at 10\% of the original class thresholds (0.9 on the x axis), still present accuracies greater than 0.97, meaning the class clusters are tightly placed near the centroids, while for digit 8, the most misclassified digit, had the threshold been set at 10\% of the original value, the accuracy would expect to drop to 0.84, meaning 16\% of the predictions would be deferred to human judgement, and 84\% would be considered 100\% accurate..

Predictions with softmax distance to class centroid below the class threshold are expected to be 100\% accurate, class predictions with distances equal or greater than threshold are then left to human judgement.



\begin{figure}[ht]
    \centering
    \includegraphics[width=0.99\columnwidth]{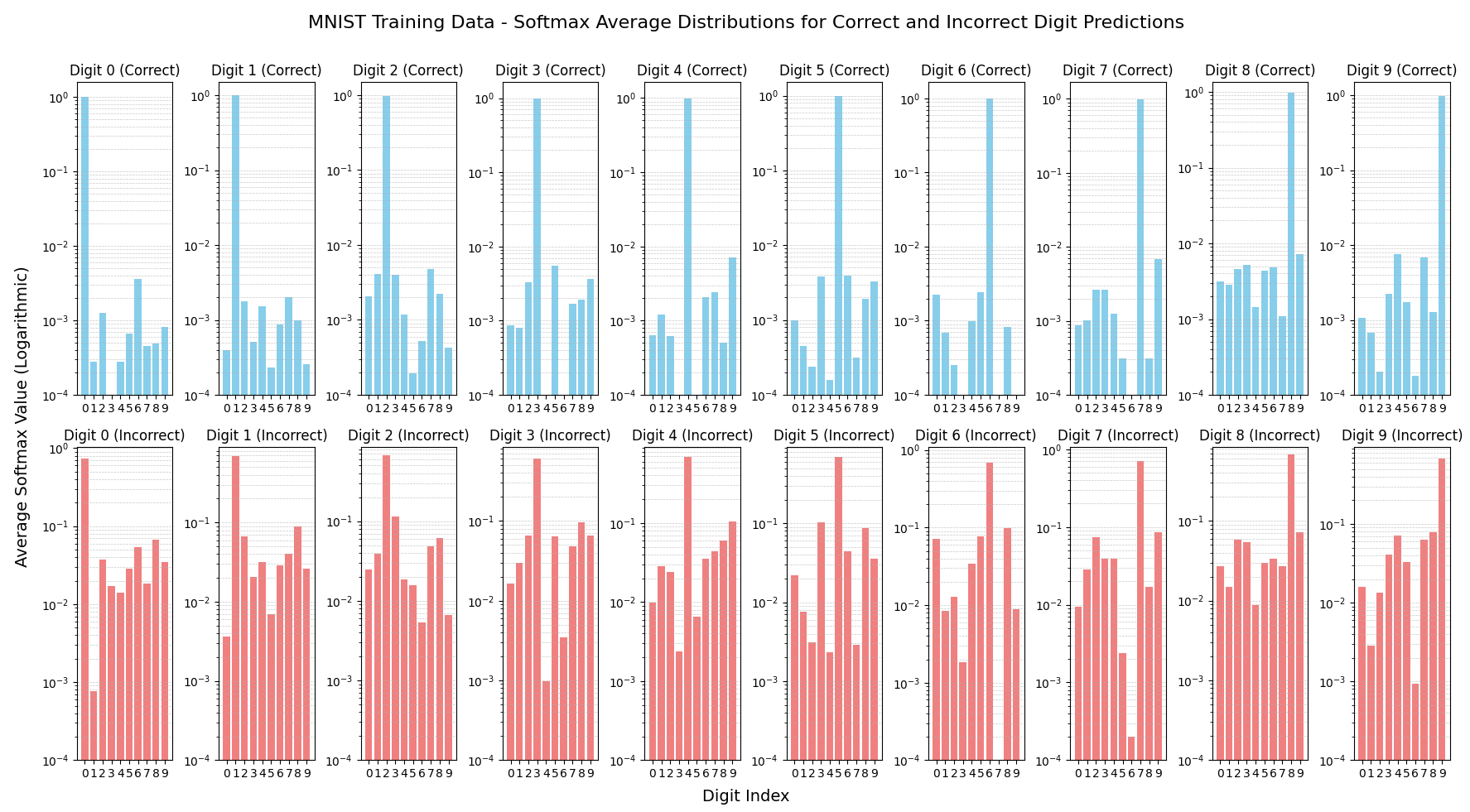}
    \caption{Please zoom in for detail. Average Softmax Probabilities for Correctly and Incorrectly Classified Digits in the MNIST Testing Dataset.}
    \label{fig:MNIST_Softmax_Averages_Training}
\end{figure}

\begin{figure}[ht]
    \centering
    \includegraphics[width=0.99\columnwidth]{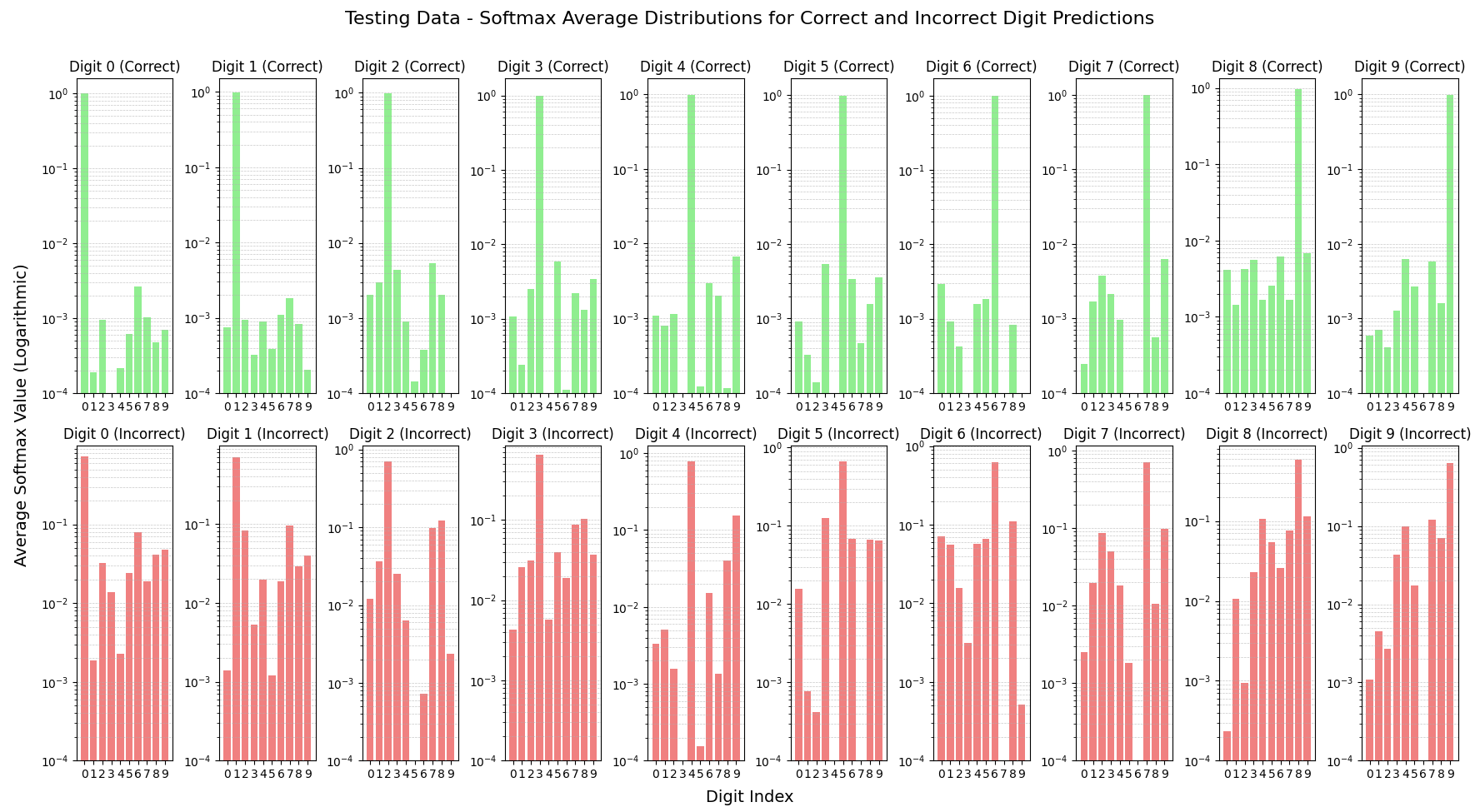}
    \caption{Please zoom in for detail. Average Softmax Probabilities for Correctly and Incorrectly Classified Digits in the MNIST Testing Dataset.}
    \label{fig:MNIST_Softmax_Averages_Testing}
\end{figure}

Figure \ref{fig:MNIST_Softmax_Averages_Training} shows the average softmax probabilities for each digit class (0 to 9) in the MNIST training dataset, separated into correctly classified instances (top row) and incorrectly classified instances (bottom row). The probabilities are displayed on a logarithmic scale. For correctly classified digits, the highest average probability is observed for the corresponding true digit class, indicating strong confidence in the correct predictions. In contrast, for incorrectly classified digits, the average probabilities are more evenly distributed across different digit classes, suggesting lower confidence and potential confusion between similar-looking digits.
Figure \ref{fig:MNIST_Softmax_Averages_Testing} is the corresponding plot for the MNIST testing dataset.

\begin{figure}[ht]
    \centering
    \includegraphics[width=0.95\columnwidth]{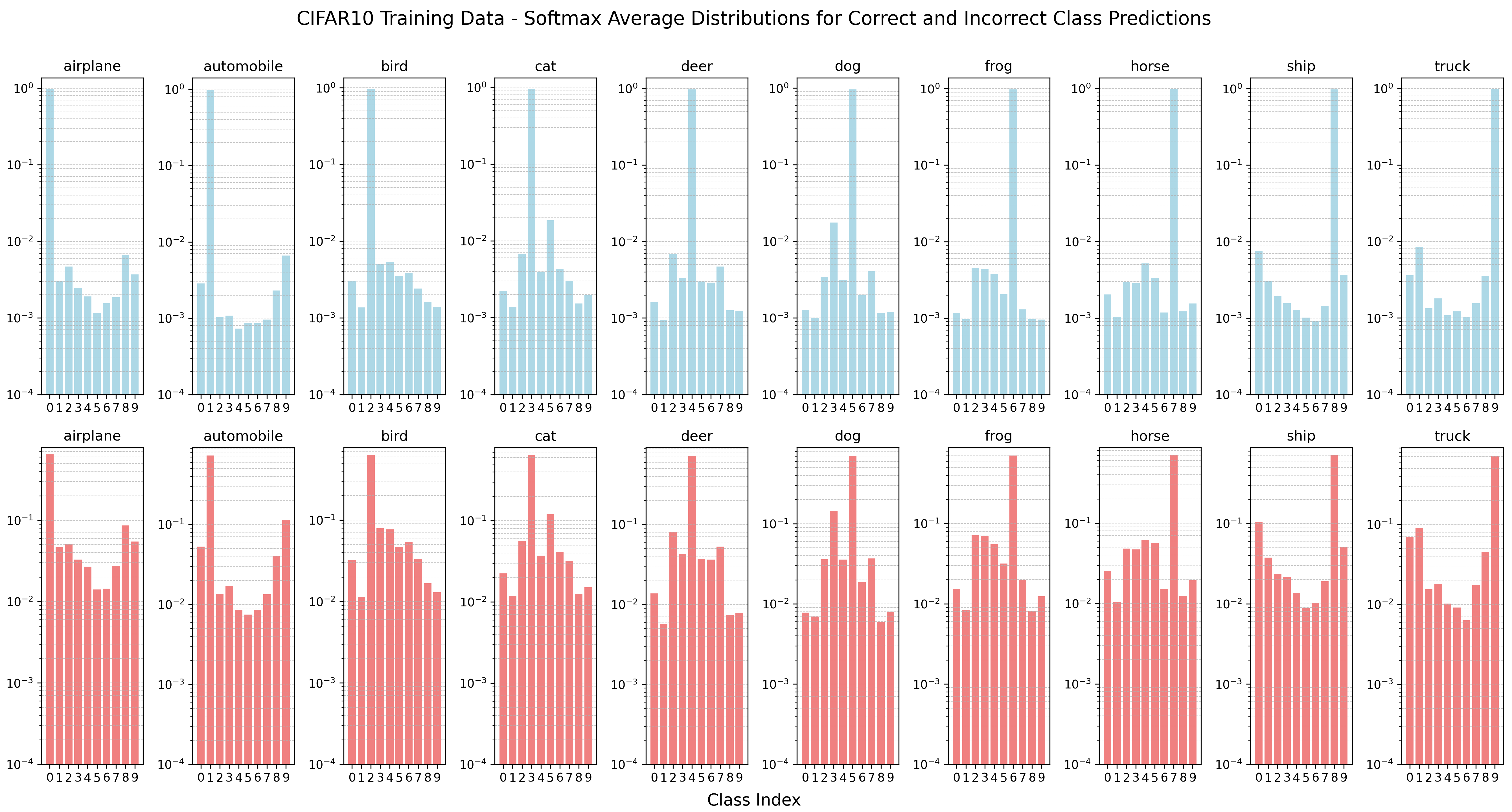}
    \caption{Please zoom in for detail. Average Softmax Probabilities for Correctly and Incorrectly Classified Classes in the CIFAR-10 Training Dataset}
    \label{fig:CIFAR10_training_plot_centroid_distance_bars.png}
\end{figure}


\begin{figure}[ht]
    \centering
    \includegraphics[width=0.95\columnwidth]{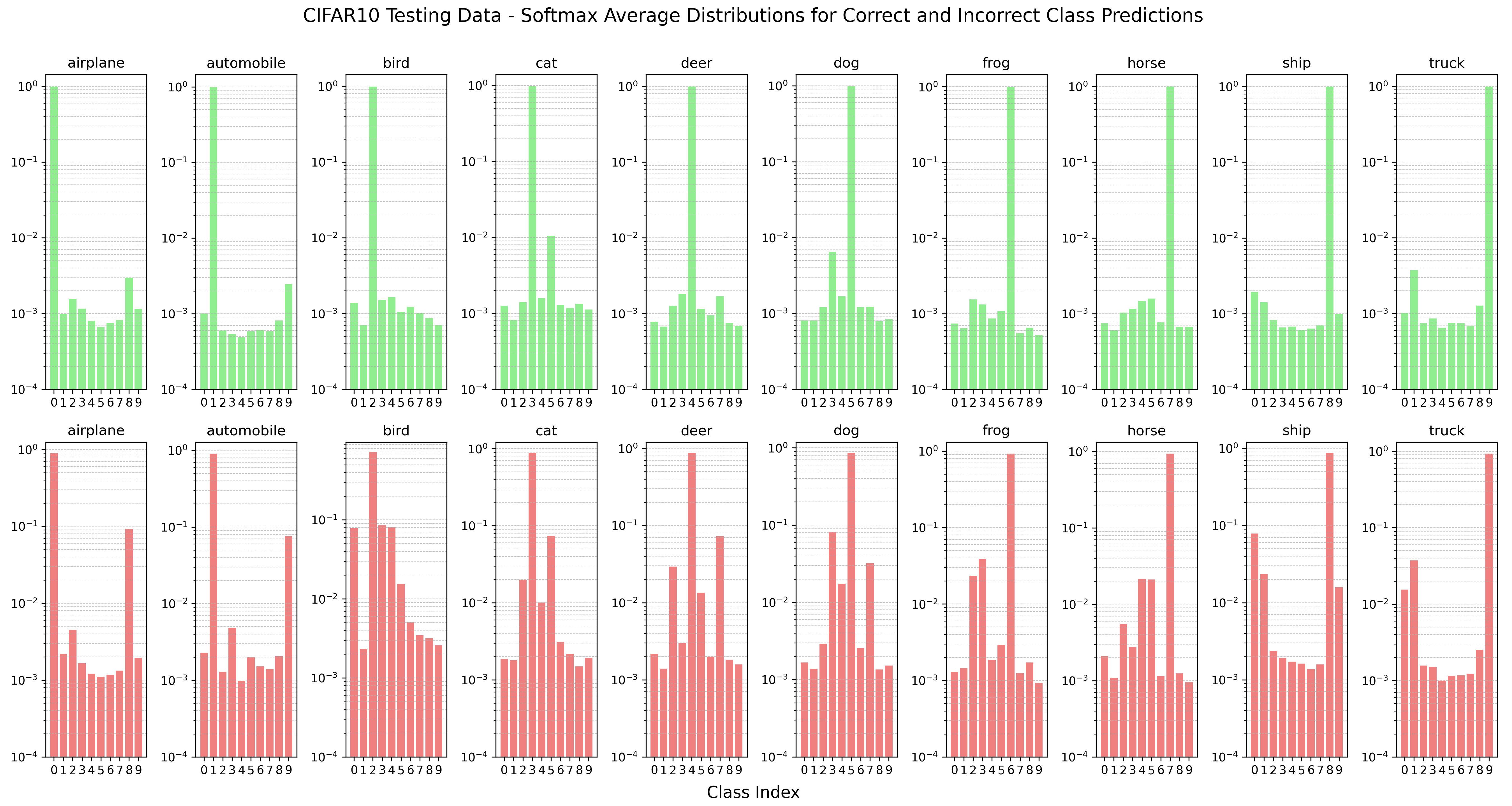}
    \caption{Please zoom in for detail. Average Softmax Probabilities for Correctly and Incorrectly Classified classes in the CIFAR-10 Testing Dataset}
    \label{fig:CIFAR10_testing_plot_centroid_distance_bars.png}
\end{figure}



For the CIFAR-10 we observe similar results. For correctly classified instances, the average softmax value corresponding to the true digit class is significantly higher compared to the softmax values of the other digit classes. This observation holds true across all digit classes in both the testing and training datasets. The pronounced difference in magnitudes indicates that the model assigns a high level of confidence to the correct predictions, with the predicted probability for the true class being orders of magnitude larger than the probabilities assigned to the other classes.

In contrast, for incorrectly classified instances, the average softmax values exhibit a more even distribution across the digit classes. While the softmax value for the predicted class is still relatively high, the softmax values for the other classes are notably closer in magnitude, as evident from the shorter bars on the logarithmic scale. This suggests that when the model makes an incorrect prediction, it assigns more substantial probabilities to multiple classes, indicating a higher level of uncertainty or confusion in the decision-making process.
Figures \ref{fig:CIFAR10_training_plot_centroid_distance_bars.png} and \ref{fig:CIFAR10_testing_plot_centroid_distance_bars.png} bar charts show the average softmax distributions for correct (green/blue) and incorrect (red) class predictions in both the testing and training CIFAR-10 datasets.

Comparing the testing and training datasets, we observe similar patterns in the softmax distance distributions for both correct and incorrect predictions. The consistency of these patterns suggests that our approach is consistent across datasets and models. 

The analysis of softmax distance distributions may be taken as an insight into the model's decision-making process. The difference in the softmax values between correct and incorrect predictions highlights the model's ability to discriminate between classes when it makes accurate predictions. On the other hand, the more evenly distributed softmax values for incorrect predictions suggest that the model struggles to make clear distinctions between classes in those cases, assigning considerable probabilities to multiple digits, where our proposed thresholding method may be an additional tool to help evaluate predictions in different settings.

%% file: Sections/06-ConclusionsAndFutureWork.tex

\section{Conclusions and Future Work}

We presented a simple approach to thresholding accuracy in an image classification setting, by training two distinct network architectures, on two distinct datasets, and showing our results are consistent across both scenarios.
We demonstrated the steps required to effectively create class clusters by initialising centroids with the mean of correct class predictions for each distinct class, where the K-Means algorithm quickly converges with good cluster separation.
A threshold was proposed to ensure predictions below threshold are safe, at a small cost of labelling as incorrect a small number of correct predictions at or above threshold.

We posit that the optimal threshold is domain-dependent and should be determined by domain experts who are equipped to balance the efficiency of automated decision-making against the necessity for time-intensive human judgment, taking into account the specific stakes involved.

We are applying the concepts discussed in this study to autonomous system safety, in the context of self-driving cars using the CARLA simulator, and constituent tasks in the decision-making stack, such as image classification and segmentation, where our methodology aims to enhance system safety by identifying scenarios, including OOD scenarios, where human judgment is preferable to the autonomous system’s assessment.


Finally, we are considering the softmax output as a training dataset in itself, for a simple multi-layer perceptron binary classifier, to compare and contrast the softmax distance approach, and a regressor network combined with our current approach, to provide a quantity e.g. sigmoid function output as a weight to further study optimal threshold values.

%% file: main.bbl
\begin{thebibliography}{52}
\providecommand{\natexlab}[1]{#1}
\providecommand{\url}[1]{\texttt{#1}}
\expandafter\ifx\csname urlstyle\endcsname\relax
  \providecommand{\doi}[1]{doi: #1}\else
  \providecommand{\doi}{doi: \begingroup \urlstyle{rm}\Url}\fi

\bibitem[Amodei et~al.(2016)Amodei, Olah, Steinhardt, Christiano, Schulman, and Man{\'e}]{amodei2016concrete}
D.~Amodei, C.~Olah, J.~Steinhardt, P.~Christiano, J.~Schulman, and D.~Man{\'e}.
\newblock Concrete problems in ai safety.
\newblock \emph{arXiv preprint arXiv:1606.06565}, 2016.

\bibitem[Arthur and Vassilvitskii(2007)]{arthur2007k}
D.~Arthur and S.~Vassilvitskii.
\newblock k-means++: The advantages of careful seeding.
\newblock In \emph{Proceedings of the eighteenth annual ACM-SIAM symposium on Discrete algorithms}, pages 1027--1035, 2007.

\bibitem[Baena-Garc{\'\i}a et~al.(2006)Baena-Garc{\'\i}a, del Campo-{\'A}vila, Fidalgo, Bifet, Gavald{\`a}, and Morales-Bueno]{baena2006early}
M.~Baena-Garc{\'\i}a, J.~del Campo-{\'A}vila, R.~Fidalgo, A.~Bifet, R.~Gavald{\`a}, and R.~Morales-Bueno.
\newblock Early drift detection method.
\newblock In \emph{Fourth International Workshop on Knowledge Discovery from Data Streams}, volume~6, pages 77--86, 2006.

\bibitem[Bishop(2006)]{bishop2006pattern}
C.~M. Bishop.
\newblock Pattern recognition and machine learning.
\newblock \emph{Springer}, 2:\penalty0 645--678, 2006.

\bibitem[Chandola et~al.(2009)Chandola, Banerjee, and Kumar]{chandola2009anomaly}
V.~Chandola, A.~Banerjee, and V.~Kumar.
\newblock Anomaly detection: A survey.
\newblock \emph{ACM computing surveys (CSUR)}, 41\penalty0 (3):\penalty0 1--58, 2009.

\bibitem[Deng et~al.(2009)Deng, Dong, Socher, Li, Li, and Fei-Fei]{deng2009imagenet}
J.~Deng, W.~Dong, R.~Socher, L.-J. Li, K.~Li, and L.~Fei-Fei.
\newblock Imagenet: A large-scale hierarchical image database.
\newblock In \emph{2009 IEEE conference on computer vision and pattern recognition}, pages 248--255. Ieee, 2009.

\bibitem[Doshi-Velez and Kim(2017)]{doshi2017towards}
F.~Doshi-Velez and B.~Kim.
\newblock Towards a rigorous science of interpretable machine learning.
\newblock \emph{arXiv preprint arXiv:1702.08608}, 2017.

\bibitem[Dosovitskiy et~al.(2020)Dosovitskiy, Beyer, Kolesnikov, Weissenborn, Zhai, Unterthiner, Dehghani, Minderer, Heigold, Gelly, et~al.]{dosovitskiy2020image}
A.~Dosovitskiy, L.~Beyer, A.~Kolesnikov, D.~Weissenborn, X.~Zhai, T.~Unterthiner, M.~Dehghani, M.~Minderer, G.~Heigold, S.~Gelly, et~al.
\newblock An image is worth 16x16 words: Transformers for image recognition at scale.
\newblock \emph{arXiv preprint arXiv:2010.11929}, 2020.

\bibitem[Eisen et~al.(1998)Eisen, Spellman, Brown, and Botstein]{eisen1998cluster}
M.~B. Eisen, P.~T. Spellman, P.~O. Brown, and D.~Botstein.
\newblock Cluster analysis and display of genome-wide expression patterns.
\newblock \emph{Proceedings of the National Academy of Sciences}, 95\penalty0 (25):\penalty0 14863--14868, 1998.

\bibitem[Ester et~al.(1996)Ester, Kriegel, Sander, and Xu]{ester1996density}
M.~Ester, H.-P. Kriegel, J.~Sander, and X.~Xu.
\newblock A density-based algorithm for discovering clusters in large spatial databases with noise.
\newblock In \emph{Kdd}, volume~96, pages 226--231, 1996.

\bibitem[Feng et~al.(2018)Feng, He, and Polson]{feng2018deep}
G.~Feng, J.~He, and N.~G. Polson.
\newblock Deep learning-based quantitative trading strategies for stock markets.
\newblock \emph{arXiv preprint arXiv:1805.01039}, 2018.

\bibitem[Gal and Ghahramani(2016)]{gal2016dropout}
Y.~Gal and Z.~Ghahramani.
\newblock Dropout as a bayesian approximation: Representing model uncertainty in deep learning.
\newblock \emph{International Conference on Machine Learning}, pages 1050--1059, 2016.

\bibitem[Gama et~al.(2014)Gama, {\v{Z}}liobait{\.e}, Bifet, Pechenizkiy, and Bouchachia]{gama2014survey}
J.~Gama, I.~{\v{Z}}liobait{\.e}, A.~Bifet, M.~Pechenizkiy, and A.~Bouchachia.
\newblock A survey on concept drift adaptation.
\newblock \emph{ACM Computing Surveys (CSUR)}, 46\penalty0 (4):\penalty0 1--37, 2014.

\bibitem[Ganin et~al.(2016)Ganin, Ustinova, Ajakan, Germain, Larochelle, Laviolette, Marchand, and Lempitsky]{ganin2016domain}
Y.~Ganin, E.~Ustinova, H.~Ajakan, P.~Germain, H.~Larochelle, F.~Laviolette, M.~Marchand, and V.~Lempitsky.
\newblock Domain-adversarial training of neural networks.
\newblock \emph{The Journal of Machine Learning Research}, 17\penalty0 (1):\penalty0 2096--2030, 2016.

\bibitem[Goodfellow et~al.(2016)Goodfellow, Bengio, and Courville]{goodfellow2016deep}
I.~Goodfellow, Y.~Bengio, and A.~Courville.
\newblock \emph{Deep learning}.
\newblock MIT press, 2016.

\bibitem[Goodfellow et~al.(2014)Goodfellow, Shlens, and Szegedy]{goodfellow2014explaining}
I.~J. Goodfellow, J.~Shlens, and C.~Szegedy.
\newblock Explaining and harnessing adversarial examples.
\newblock In \emph{International Conference on Learning Representations}, 2014.

\bibitem[Gretton et~al.(2012)Gretton, Borgwardt, Rasch, Sch{\"o}lkopf, and Smola]{gretton2012kernel}
A.~Gretton, K.~M. Borgwardt, M.~J. Rasch, B.~Sch{\"o}lkopf, and A.~Smola.
\newblock A kernel two-sample test.
\newblock \emph{The Journal of Machine Learning Research}, 13\penalty0 (1):\penalty0 723--773, 2012.

\bibitem[Guo et~al.(2017)Guo, Pleiss, Sun, and Weinberger]{guo2017calibration}
C.~Guo, G.~Pleiss, Y.~Sun, and K.~Q. Weinberger.
\newblock On calibration of modern neural networks.
\newblock In \emph{International Conference on Machine Learning}, pages 1321--1330. PMLR, 2017.

\bibitem[Guo et~al.(2018)Guo, Rana, Cisse, and Van Der~Maaten]{guo2018countering}
C.~Guo, M.~Rana, M.~Cisse, and L.~Van Der~Maaten.
\newblock Countering adversarial images using input transformations.
\newblock In \emph{International Conference on Learning Representations}, 2018.

\bibitem[Hendrycks and Dietterich(2019)]{hendrycks2019benchmarking}
D.~Hendrycks and T.~Dietterich.
\newblock Benchmarking neural network robustness to common corruptions and perturbations.
\newblock \emph{arXiv preprint arXiv:1903.12261}, 2019.

\bibitem[Hendrycks and Gimpel(2017)]{hendrycks17baseline}
D.~Hendrycks and K.~Gimpel.
\newblock A baseline for detecting misclassified and out-of-distribution examples in neural networks.
\newblock In \emph{International Conference on Learning Representations}, 2017.

\bibitem[Hendrycks et~al.(2021)Hendrycks, Basart, Mu, Kadavath, Wang, Dorundo, Desai, Zhu, Parajuli, Guo, et~al.]{hendrycks2021many}
D.~Hendrycks, S.~Basart, N.~Mu, S.~Kadavath, F.~Wang, E.~Dorundo, R.~Desai, T.~Zhu, S.~Parajuli, M.~Guo, et~al.
\newblock The many faces of robustness: A critical analysis of out-of-distribution generalization.
\newblock In \emph{Proceedings of the IEEE/CVF international conference on computer vision}, pages 8340--8349, 2021.

\bibitem[Jain(2010)]{jain2010data}
A.~K. Jain.
\newblock Data clustering: 50 years beyond k-means.
\newblock \emph{Pattern recognition letters}, 31\penalty0 (8):\penalty0 651--666, 2010.

\bibitem[Jiang et~al.(2004)Jiang, Tang, and Zhang]{jiang2004cluster}
D.~Jiang, C.~Tang, and A.~Zhang.
\newblock Cluster analysis for gene expression data: A survey.
\newblock \emph{IEEE Transactions on knowledge and data engineering}, 16\penalty0 (11):\penalty0 1370--1386, 2004.

\bibitem[Johnson(1967)]{johnson1967hierarchical}
S.~C. Johnson.
\newblock Hierarchical clustering schemes.
\newblock \emph{Psychometrika}, 32\penalty0 (3):\penalty0 241--254, 1967.

\bibitem[Kendall and Gal(2017)]{kendall2017uncertainties}
A.~Kendall and Y.~Gal.
\newblock What uncertainties do we need in bayesian deep learning for computer vision?
\newblock \emph{Advances in Neural Information Processing Systems}, 30, 2017.

\bibitem[Kullback and Leibler(1951)]{kullback1951information}
S.~Kullback and R.~A. Leibler.
\newblock On information and sufficiency.
\newblock \emph{The Annals of Mathematical Statistics}, 22\penalty0 (1):\penalty0 79--86, 1951.

\bibitem[Lakshminarayanan et~al.(2017)Lakshminarayanan, Pritzel, and Blundell]{lakshminarayanan2017simple}
B.~Lakshminarayanan, A.~Pritzel, and C.~Blundell.
\newblock Simple and scalable predictive uncertainty estimation using deep ensembles.
\newblock In \emph{Advances in Neural Information Processing Systems}, pages 6402--6413, 2017.

\bibitem[Leibig et~al.(2017)Leibig, Allken, Ayhan, Berens, and Wahl]{leibig2017leveraging}
C.~Leibig, V.~Allken, M.~S. Ayhan, P.~Berens, and S.~Wahl.
\newblock Leveraging uncertainty information from deep neural networks for disease detection.
\newblock \emph{Scientific Reports}, 7\penalty0 (1):\penalty0 1--14, 2017.

\bibitem[Liang et~al.(2018)Liang, Li, and Srikant]{liang2018enhancing}
S.~Liang, Y.~Li, and R.~Srikant.
\newblock Enhancing the reliability of out-of-distribution image detection in neural networks.
\newblock In \emph{International Conference on Learning Representations}, 2018.

\bibitem[Lloyd(1982)]{lloyd1982least}
S.~Lloyd.
\newblock Least squares quantization in pcm.
\newblock \emph{IEEE transactions on information theory}, 28\penalty0 (2):\penalty0 129--137, 1982.

\bibitem[Lu et~al.(2018)Lu, Liu, Dong, Gu, Gama, and Zhang]{lu2018learning}
J.~Lu, A.~Liu, F.~Dong, F.~Gu, J.~Gama, and G.~Zhang.
\newblock Learning under concept drift: A review.
\newblock \emph{IEEE Transactions on Knowledge and Data Engineering}, 31\penalty0 (12):\penalty0 2346--2363, 2018.

\bibitem[Madry et~al.(2017)Madry, Makelov, Schmidt, Tsipras, and Vladu]{madry2017towards}
A.~Madry, A.~Makelov, L.~Schmidt, D.~Tsipras, and A.~Vladu.
\newblock Towards deep learning models resistant to adversarial attacks.
\newblock In \emph{International Conference on Learning Representations}, 2017.

\bibitem[Michelmore et~al.(2018)Michelmore, Kwiatkowska, and Gal]{michelmore2018evaluating}
R.~Michelmore, M.~Kwiatkowska, and Y.~Gal.
\newblock Evaluating uncertainty quantification in end-to-end autonomous driving control.
\newblock \emph{arXiv preprint arXiv:1811.06817}, 2018.

\bibitem[Moreno-Torres et~al.(2012)Moreno-Torres, Raeder, Alaiz-Rodr{\'\i}guez, Chawla, and Herrera]{moreno2012unifying}
J.~G. Moreno-Torres, T.~Raeder, R.~Alaiz-Rodr{\'\i}guez, N.~V. Chawla, and F.~Herrera.
\newblock A unifying view on dataset shift in classification.
\newblock \emph{Pattern Recognition}, 45\penalty0 (1):\penalty0 521--530, 2012.

\bibitem[M{\"u}llner(2011)]{mullner2011modern}
D.~M{\"u}llner.
\newblock Modern hierarchical, agglomerative clustering algorithms.
\newblock \emph{arXiv preprint arXiv:1109.2378}, 2011.

\bibitem[Ngai et~al.(2009)Ngai, Xiu, and Chau]{ngai2009application}
E.~W. Ngai, L.~Xiu, and D.~C. Chau.
\newblock Application of data mining techniques in customer relationship management: A literature review and classification.
\newblock \emph{Expert systems with applications}, 36\penalty0 (2):\penalty0 2592--2602, 2009.

\bibitem[Pan and Yang(2009)]{pan2009survey}
S.~J. Pan and Q.~Yang.
\newblock A survey on transfer learning.
\newblock \emph{IEEE Transactions on Knowledge and Data Engineering}, 22\penalty0 (10):\penalty0 1345--1359, 2009.

\bibitem[Patel et~al.(2015)Patel, Gopalan, Li, and Chellappa]{patel2015visual}
V.~M. Patel, R.~Gopalan, R.~Li, and R.~Chellappa.
\newblock Visual domain adaptation: A survey of recent advances.
\newblock \emph{IEEE Signal Processing Magazine}, 32\penalty0 (3):\penalty0 53--69, 2015.

\bibitem[Qui{\~n}onero-Candela et~al.(2009)Qui{\~n}onero-Candela, Sugiyama, Schwaighofer, and Lawrence]{quinonero2009dataset}
J.~Qui{\~n}onero-Candela, M.~Sugiyama, A.~Schwaighofer, and N.~D. Lawrence.
\newblock Dataset shift in machine learning.
\newblock \emph{The MIT Press}, 2009.

\bibitem[Ribeiro et~al.(2016)Ribeiro, Singh, and Guestrin]{ribeiro2016should}
M.~T. Ribeiro, S.~Singh, and C.~Guestrin.
\newblock "why should i trust you?" explaining the predictions of any classifier.
\newblock In \emph{Proceedings of the 22nd ACM SIGKDD International Conference on Knowledge Discovery and Data Mining}, pages 1135--1144, 2016.

\bibitem[Rodriguez et~al.(2019)Rodriguez, Comin, Casanova, Bruno, Amancio, Costa, and Rodrigues]{rodriguez2019clustering}
M.~Z. Rodriguez, C.~H. Comin, D.~Casanova, O.~M. Bruno, D.~R. Amancio, L.~d.~F. Costa, and F.~A. Rodrigues.
\newblock Clustering algorithms: A comparative approach.
\newblock \emph{PloS one}, 14\penalty0 (1):\penalty0 e0210236, 2019.

\bibitem[Schubert et~al.(2017)Schubert, Sander, Ester, Kriegel, and Xu]{schubert2017dbscan}
E.~Schubert, J.~Sander, M.~Ester, H.~P. Kriegel, and X.~Xu.
\newblock Dbscan revisited, revisited: why and how you should (still) use dbscan.
\newblock \emph{ACM Transactions on Database Systems (TODS)}, 42\penalty0 (3):\penalty0 1--21, 2017.

\bibitem[Shi and Malik(2000)]{shi2000normalized}
J.~Shi and J.~Malik.
\newblock Normalized cuts and image segmentation.
\newblock \emph{IEEE Transactions on pattern analysis and machine intelligence}, 22\penalty0 (8):\penalty0 888--905, 2000.

\bibitem[Shimodaira(2000)]{shimodaira2000improving}
H.~Shimodaira.
\newblock Improving predictive inference under covariate shift by weighting the log-likelihood function.
\newblock \emph{Journal of Statistical Planning and Inference}, 90\penalty0 (2):\penalty0 227--244, 2000.

\bibitem[Sugiyama et~al.(2007)Sugiyama, Krauledat, and M{\"u}ller]{sugiyama2007covariate}
M.~Sugiyama, M.~Krauledat, and K.-R. M{\"u}ller.
\newblock Covariate shift adaptation by importance weighted cross validation.
\newblock \emph{The Journal of Machine Learning Research}, 8:\penalty0 985--1005, 2007.

\bibitem[Szegedy et~al.(2013)Szegedy, Zaremba, Sutskever, Bruna, Erhan, Goodfellow, and Fergus]{szegedy2013intriguing}
C.~Szegedy, W.~Zaremba, I.~Sutskever, J.~Bruna, D.~Erhan, I.~Goodfellow, and R.~Fergus.
\newblock Intriguing properties of neural networks.
\newblock In \emph{International Conference on Learning Representations}, 2013.

\bibitem[Wang and Deng(2018)]{wang2018deep}
M.~Wang and W.~Deng.
\newblock Deep visual domain adaptation: A survey.
\newblock \emph{Neurocomputing}, 312:\penalty0 135--153, 2018.

\bibitem[Wolf et~al.(2020)Wolf, Debut, Sanh, Chaumond, Delangue, Moi, Cistac, Rault, Louf, Funtowicz, Davison, Shleifer, von Platen, Ma, Jernite, Plu, Xu, Scao, Gugger, Drame, Lhoest, and Rush]{wolf2020huggingfaces}
T.~Wolf, L.~Debut, V.~Sanh, J.~Chaumond, C.~Delangue, A.~Moi, P.~Cistac, T.~Rault, R.~Louf, M.~Funtowicz, J.~Davison, S.~Shleifer, P.~von Platen, C.~Ma, Y.~Jernite, J.~Plu, C.~Xu, T.~L. Scao, S.~Gugger, M.~Drame, Q.~Lhoest, and A.~M. Rush.
\newblock Huggingface's transformers: State-of-the-art natural language processing, 2020.

\bibitem[Wu et~al.(2020)Wu, Xu, Dai, Wan, Zhang, Yan, Tomizuka, Gonzalez, Keutzer, and Vajda]{wu2020visual}
B.~Wu, C.~Xu, X.~Dai, A.~Wan, P.~Zhang, Z.~Yan, M.~Tomizuka, J.~Gonzalez, K.~Keutzer, and P.~Vajda.
\newblock Visual transformers: Token-based image representation and processing for computer vision, 2020.

\bibitem[Xu and Tian(2015)]{xu2015comprehensive}
D.~Xu and Y.~Tian.
\newblock A comprehensive survey of clustering algorithms.
\newblock \emph{Annals of Data Science}, 2\penalty0 (2):\penalty0 165--193, 2015.

\bibitem[Zadrozny and Elkan(2002)]{zadrozny2002transforming}
B.~Zadrozny and C.~Elkan.
\newblock Transforming classifier scores into accurate multiclass probability estimates.
\newblock In \emph{Proceedings of the eighth ACM SIGKDD International Conference on Knowledge Discovery and Data Mining}, pages 694--699, 2002.

\end{thebibliography}
